\documentclass[letterpaper]{article}
\usepackage{aaai,times,helvet,courier}
\usepackage[protrusion=true,spacing=true]{microtype}
\usepackage[group-separator={,}]{siunitx}
\usepackage{algorithmic,algorithm}
\usepackage{amsmath,amsfonts}
\usepackage{graphicx,subfig}
\usepackage[absolute,overlay]{textpos}
\newcommand{\ehba}{\text{E-HBA}}
\newcommand{\pr}{\text{Pr}}
\newcommand{\y}{\fontsize{11}{13}\selectfont}
\newcommand{\x}{\fontsize{10}{12}\selectfont\textnormal}
\newcommand{\cit}[1]{\citeauthor{#1} \citeyear{#1}}
\newcommand{\citp}[1]{\citeauthor{#1} \shortcite{#1}}
\newcommand{\h}{0.186\textheight}
\newcommand{\hs}{15pt}

\begin{document}

	\title{E-HBA: Using Action Policies for Expert Advice and Agent Typification \\[2pt]}

	\author{
		\begin{tabular}{@{}ccc@{}}
			\y{Stefano V. Albrecht} & \y{Jacob W. Crandall} & \y{Subramanian Ramamoorthy}\hspace{-1pt} \\[-2pt]
			\x{The University of Edinburgh} & \x{Masdar Institute of Science and Technology} & \x{The University of Edinburgh} \\[-2pt]
			\x{Edinburgh, United Kingdom} & \x{Abu Dhabi, United Arab Emirates} & \x{Edinburgh, United Kingdom} \\[-2.5pt]
			\x{s.v.albrecht@sms.ed.ac.uk} & \x{jcrandall@masdar.ac.ae} & \x{s.ramamoorthy@ed.ac.uk}
		\end{tabular}
	}

	\maketitle

	\begin{abstract}
		\begin{quote}
			Past research has studied two approaches to utilise pre-defined policy sets in repeated interactions: as \emph{experts}, to dictate our own actions, and as \emph{types}, to characterise the behaviour of other agents. In this work, we bring these complementary views together in the form of a novel meta-algorithm, called \emph{Expert-HBA} (\ehba), which can be applied to any expert algorithm that considers the average (or total) payoff an expert has yielded in the past. \ehba\ gradually mixes the past payoff with a predicted future payoff, which is computed using the type-based characterisation. We present results from a comprehensive set of repeated matrix games, comparing the performance of several well-known expert algorithms with and without the aid of \ehba. Our results show that \ehba\ has the potential to significantly improve the performance of expert algorithms.
		\end{quote}
	\end{abstract}

	\section{Introduction} \label{sec:intro}

Many multiagent applications require an agent to quickly learn how to interact effectively with previously unknown other agents. Important examples include electronic markets, adaptive user interfaces, and robotic elderly care. Learning effective policies from scratch in such applications (e.g. using reinforcement learning or opponent modelling) can be difficult because of the essentially unconstrained nature of the interaction problem, by which we mean that the other agents may, in principle, have any kind of behaviours.

One approach to make this problem more tractable is to assume that we have access to a set of policies, or \emph{experts}, which make action recommendations based on the current interaction history. Such experts may be specified by a human user or generated automatically from the problem specification. The goal, then, is to find the best expert through some repeated exploration strategy. Several such strategies, or \emph{expert algorithms}, have been defined, e.g. \cite{c2014,fm2004,acfs1995}.

Another approach is to use such policies, then called \emph{types}, to characterise the behaviour of other agents. In this approach, the observed actions of an agent are compared with the predictions of the types, resulting in a posterior distribution which describes the relative likelihood that the agent implements any of the types. This distribution can then be used in a planning procedure to compute best-response actions, e.g. \cite{ar2014,bsk2011,cm1999}.

While both approaches have been shown to be effective under various circumstances, they have certain limitations: Most expert algorithms are \emph{reactive} in the sense that they rely heavily on the past performance of experts, typically in the form of the average payoff an expert yielded. However, this can be problematic when interacting with adaptive agents, in which case past performance is not necessarily a good indicator of future performance. On the other hand, type-based methods such as HBA \cite{ar2014} are \emph{proactive} in that they make explicit predictions about future interactions and choose actions accordingly. However, such methods are not designed for situations in which none of the types account for the observed behaviour of an agent (i.e. all posterior probabilities are zero or undefined).

To illustrate these limitations, suppose we are playing the Prisoner's Dilemma game against a simple adaptive opponent that cooperates only if we cooperated in the past 4 rounds, otherwise it defects. Assume we are given two experts, C~and~D, where C always cooperates and D always defects. A typical expert algorithm would try both experts and may find that D has a higher average payoff than C, if C is not tried for a sufficient number of consecutive rounds. Thus, it would favour D, which in the long-run is the worse expert. On the other hand, if the opponent's behaviour was provided to HBA as a type, it would eventually learn the true type and play optimally against it (that is, provided that its planning horizon is deep enough). However, if the true type is unknown to HBA, then the posterior may become undefined in the worst case and HBA would play randomly.

In this paper, we propose to address these limitations by combining the two approaches. Specifically, we present a novel meta-algorithm, called \emph{Expert-HBA} (\ehba), which can be applied to any expert algorithm that considers the average (or total) payoffs the experts yielded when following their recommendations. \ehba\ gradually mixes the past payoff an expert yielded with a predicted future payoff, which is computed using the type-based approach. The mixing is gradual in that the weight of the predicted payoff is proportional to the \emph{confidence} \ehba\ has in the correctness of its predictions (to be made precise shortly). We present results from a comprehensive set of repeated matrix games, comparing the performance of several well-known expert algorithms with and without the aid of \ehba. Our results show that, if the true (or a similar) type of the opponent is in \ehba's set of types, then it can significantly improve the performance of the expert algorithm, while in all other cases it performs similarly to the original expert algorithm.

	\section{Related Work} \label{sec:relwork}

A number of expert algorithms have been defined. Some of the more well-known ones include the 'Weighted Majorities' algorithm \cite{lw1994}, Hedge \cite{fs1995}, the Exp-family \cite{acfs1995}, the UCB-family \cite{acf2002}, EEE \cite{fm2004}, and S \cite{kmrv1998}.

\ehba\ itself is not an expert algorithm. Rather, it is a meta-algorithm that can be applied to any expert algorithm which considers the average (or total) payoff an expert has yielded. This includes all of the above algorithms, to which \ehba\ can be applied in a straight-forward manner.

Our work is closest in spirit to \cite{c2014}, who proposed a meta-algorithm that prunes the set of selectable experts to those which it considers most promising. This is done using a variant of aspiration learning, similar to \cite{kmrv1998}. \ehba\ can be combined with Crandall's meta-algorithm since they target different aspects in expert algorithms: \ehba\ modifies the average payoffs of experts while Crandall's algorithm prunes the set of experts.

Most expert algorithms are evaluated in terms of \emph{regret} (e.g. \cit{fv1999}), which is the difference between the received payoffs and the payoffs of the best expert against the observed actions. However, it has been argued that this view of regret may be inadequate in interactive settings with adaptive agents \cite{c2014,adt2012}. In this work, we evaluate expert algorithms in terms of the average payoffs they achieved.

\ehba\ is based on HBA \cite{ar2014,ar2013}, which is informally described in Section~\ref{sec:intro} and formally defined in Section~\ref{sec:hba}. Related methods were studied by \citp{bsk2011} and \citp{cm1999}. We focus on HBA because it is a relatively simple and general method with well-known guarantees \cite{ar2014}.

The methods used in I-POMDPs \cite{gd2005} are closely related to HBA. However, as they are designed to handle the full generality of partially observable states, complex nested beliefs, and subjective optimality criteria, their solutions can be very hard to compute. We focus on HBA because we are not aware of any expert algorithm that was specifically designed to handle partially observable states. However, we believe that our work can be extended to more complex models such as I-POMDPs.

	\section{Preliminaries} \label{sec:prel}

This section introduces some basic notation and definitions which we will use in the remainder of the paper.

		\subsection{Model}

To simplify the exposition, we focus on $2$-player repeated matrix games. However, our definitions can be extended to more complex models such as stochastic Bayesian games \cite{ar2014} and I-POMDPs \cite{gd2005}.

We use $i$ to refer to our player and $j$ to refer to the other player. At each time $t$ in the game, each player $k \in \left\{ i,j \right\}$ independently chooses an action $a_k^t \in A_k$ and receives a payoff $u_k(a^t_i,a^t_j) \in \mathbb{R}$. This process is repeated for a finite number of rounds. We assume we know $A_i$, $A_j$, and $u_i$.

		\subsection{Expert Algorithms} \label{sec:expalg}

Let $\Phi_i$ be a set of \emph{experts} for player $i$. Each expert $\phi_i \in \Phi_i$ specifies an action policy for player $i$. We write $\pi_i(H^t,a_i,\phi_i)$ to denote the probability that expert $\vspace{1pt}\phi_i$ chooses action $a_i$ after the history $\vspace{-1pt}H^t = \langle (a_i^0,a_j^0),...,(a_i^{t-1},a_j^{t-1}) \rangle$ of previous joint actions. Furthermore, we use $\bar{U}$ to denote the vector of average payoffs the experts have yielded when following their recommendations (i.e. one entry for each expert).

For the purposes of this exposition, we abstractly define an \emph{expert algorithm} as a function $f(\bar{U})$ which returns a probability distribution over the set $\Phi_i$ of experts. This distribution specifies what experts to choose at any given time. We note that many expert algorithms use auxiliary statistics, such as the number of times an expert has been chosen. However, in the interest of clarity, we omit such details here.

		\subsection{Harsanyi-Bellman Ad Hoc Coordination} \label{sec:hba}

Our meta-algorithm is based on \emph{Harsanyi-Bellman Ad Hoc Coordination} (HBA) \cite{ar2014}. HBA utilises a set $\Theta_j^*$ of hypothesised \emph{types} which specify complete action policies for player $j$. The types are hypothetical in the sense that player $j$ may or may not implement one of the types in $\Theta_j^*$, but we do not a priori know this.

We write $\pi_j(H^t,a_j,\theta_j^*)$ to denote the probability that player $j$ chooses action $a_j$ if it is of type $\theta_j^*$, given the history $H^t$ of previous joint actions. HBA computes the posterior belief that player $j$ is of type $\theta_j^*$, given $H^t$, as
\begin{equation}\label{eq:pos}
	\pr_j(\theta_j^* | H^t) = \eta \, P_j(\theta_j^*) \prod_{\tau=0}^{t-1} \pi_j(H^{\tau}\hspace{-1pt},a_j^\tau,\theta_j^*)
\end{equation}
where $\eta$ is a normalisation constant and $P_j(\theta_j^*)$ denotes the prior belief (e.g. uniform) that player $j$ is of type $\theta_j^*$.

Using the posterior $\pr_j$, HBA chooses an action $a_i$ which maximises the expected average payoff $\frac{1}{h}E_h^{a_i}(H^t)$, where
\vspace{5pt}
\begin{equation}\label{eq:EQ}
	E_h^{a_i}(\hat{H}) = \hspace{-4pt} \sum_{\theta_j^* \in \Theta_j^*} \hspace{-3pt} \pr_j(\theta_j^* | H^t) \hspace{-2pt} \sum_{a_j \in A_j} \hspace{-3pt} \pi_j(\hat{H},a_j,\theta_j^*) \, Q_{h-1}^{(a_i,a_j)}(\hat{H})
\end{equation}
\begin{equation*}
	Q_h^{(a_i,a_j)}(\hat{H}) = u_i(a_i,a_j) + \left\{ \begin{array}{l} 0 \ \ \text{if}\ \ h = 0, \ \text{else} \\ \max_{a'_i} E_h^{a'_i} \hspace{-2pt} \left( \langle \hat{H},(a_i,a_j)\rangle \right) \end{array} \right.\vspace{5pt}
\end{equation*}
and $h$ specifies the depth of the planning horizon. Note that $H^t$ is the current history while $\hat{H}$ is used to construct all future trajectories in the game.

	\section{Expert-HBA} \label{sec:ehba}

\emph{Expert-HBA} (\ehba) is a meta-algorithm that can be applied to any expert algorithm which is in the general form of $f(\bar{U})$. A schematic of \ehba\ is provided in Algorithm~\ref{alg:ehba}.

		\subsection{Future Payoffs of Experts}

\ehba\ mixes the vector $\bar{U}$ of \emph{observed} (past) average payoffs of each expert with a vector $U^*$ of \emph{expected} (future) average payoffs for each expert. In order to compute $U^*$, \ehba\ uses an adapted version of HBA.

Similar to HBA, \ehba\ maintains a posterior distribution $\pr_j$ over a set $\Theta_j^*$ of hypothesised types for player $j$, which can be obtained in the same way as the expert set $\Phi_i$. This distribution is computed using \eqref{eq:pos}. To obtain the vector $U^*$, \ehba\ computes the expected average payoff of each expert $\phi_i \in \Phi_i$, using a modified version of \eqref{eq:EQ} which redefines
\begin{equation}\label{eq:Qmod}
	\begin{array}{l} Q_h^{(a_i,a_j)} = \\ u_i(a_i,a_j) + \left\{ \begin{array}{l} 0 \ \ \text{if}\ \ h = 0, \ \text{else} \\ \sum_{a'_i} \pi_i(\hat{H},a'_i,\phi_i) \, E_h^{a'_i} \hspace{-2pt} \left( \langle \hat{H},(a_i,a_j)\rangle \right) \end{array} \right.\hspace{-6pt}. \end{array}
\end{equation}

The difference between \eqref{eq:Qmod} and the original definition is that \eqref{eq:Qmod} only follows the expert when expanding future trajectories in the game, whereas the original definition has to consider all possible trajectories in order to implement the $\max$-operator. Depending on the stochasticity of the experts (i.e. the number of actions with positive probability), this may mean that \eqref{eq:Qmod} can be computed more efficiently than the original definition. Note that, as with the original definition, \eqref{eq:Qmod} may be approximated efficiently using Monte-Carlo Tree Search, e.g. \cite{ks2006}.

		\subsection{Mixing with Confidence}

Given the vectors $\bar{U}$ and $U^*$, \ehba\ executes the expert algorithm as $f \hspace{-2pt}\left( (1-C^t) \bar{U} + C^t U^* \right)$, where $C^t \in [0,1]$ is the key mixing factor that gradually combines the expert and type methodologies. Intuitively, $C^t$ can be interpreted as the \emph{confidence} \ehba\ has at time $t$ in the correctness of $U^*$, such that $C^t = 1$ corresponds to absolute confidence and $C^t = 0$ corresponds to no confidence at all.

Let the true type of player $j$ be denoted by $\theta_j^+$, and assume for simplicity that we have a single hypothesised type $\theta_j^*$ (any pair $(\pr_j,\Theta_j^*)$ can be represented as a single type). Then, $C^t$ can be viewed as quantifying the \emph{similarity} between $\theta_j^+$ and $\theta_j^*$. However, given that we only observe $H^t$ and $\theta_j^*$ but not $\theta_j^+$, this can be an extremely difficult task. Indeed, even if we knew the true type $\theta_j^+$, it would by no means be clear how to best quantify a similarity between $\theta_j^+$ and $\theta_j^*$.

In this preliminary work, we define confidence as the average weighted ratio of probabilities assigned to observed actions and maximum probabilities prescribed by the types, where the weight is given by the posterior $\pr_j$. Formally, for $t > 0$,
\begin{equation}\label{eq:C}
	C^t = \frac{1}{t} \sum_{\tau = 0}^{t-1} \sum_{\, \theta_j^* \in \,\Theta_j^*} \hspace{-3pt} \pr_j(\theta_j^* | H^{\tau}) \, \frac{\pi_j(H^{\tau}\hspace{-1pt},a_j^{\tau},\theta_j^*)}{\max_{a_j} \pi_j(H^{\tau}\hspace{-1pt},a_j,\theta_j^*)}
\end{equation}
where $C^0$ can be set arbitrarily in $[0,1]$, e.g. $C^0 = 1$ to indicate extreme optimism and $C^0 = 0$ for extreme pessimism. However, $C^0$ has typically no bearing because most expert algorithms choose the first expert randomly.

This definition of confidence is a useful baseline because it often approximates the average probability overlap between $\theta_j^*$ and $\theta_j^+$, which is one possible quantification of similarity. Again assuming a single hypothesised type $\theta_j^*$, the average probability overlap between $\theta_j^*$ and $\theta_j^+$ at time $t > 0$ is
\begin{equation*}
	\frac{1}{t}\ \sum_{\tau = 0}^{t-1} \sum_{\ a_j \in A_j} \hspace{-2pt} \min \left[ \pi_j(H^\tau,a_j,\theta_j^*) , \pi_j(H^\tau,a_j,\theta_j^+) \right].
\end{equation*}
However, we do note that \eqref{eq:C} is not a perfect choice due to several shortcomings. For instance, \eqref{eq:C} will always converge to 1 if $\theta_j^*$ converges to uniform action probabilities. Thus, more research is required to formulate a suitable theory around this notion of confidence.


\begin{algorithm}[t]
	\begin{algorithmic}
		\STATE Let $\bar{U}$ be vector of observed average payoffs of experts\\[1pt]
		\STATE Compute posterior $\pr_j$ using \eqref{eq:pos}\\[1pt]
		\STATE Compute vector $U^*$ using \eqref{eq:EQ}/\eqref{eq:Qmod}\\[1pt]
		\STATE Compute confidence $C^t$, e.g. using \eqref{eq:C}\\[1pt]
		\STATE Execute expert algorithm $f \hspace{-2pt}\left( (1-C^t) \bar{U} + C^t U^* \right)$
		\vspace{-2pt}
	\end{algorithmic}
	\caption{Schematic of Expert-HBA (\ehba)}
	\label{alg:ehba}
\end{algorithm}

	\subsection{Guarantees} \label{sec:guaran}

Since \ehba\ maintains the posterior $\pr_j$ in the same way as HBA, it inherits all convergence guarantees of HBA. This includes Theorems 1 to 3 in \cite{ar2014}. For example, Theorem 1 states that, if the true type of player $j$ is included in $\Theta_j^*$ and if the prior beliefs $P_j$ are positive (i.e. $P_j(\theta_j^*) > 0$ for all $\theta_j^* \in \Theta_j^*$), then \ehba's predictions of future play will eventually be correct.

This reveals an interesting property of \ehba: Once \ehba\ makes correct future predictions, and for sufficiently high $h$, the predicted payoffs $U^*$ will be the \emph{true} average payoffs of the experts. Therefore, mixing $\bar{U}$ and $U^*$ will be \emph{more accurate} than $\bar{U}$ alone (unless $\bar{U} = U^*$), for any $C^t > 0$. Thus, under these circumstances, the mixing will not degrade the performance of the expert algorithm.

Of course, whether $h$ is sufficient to produce accurate predictions $U^*$ depends on the types in $\Theta_j^*$. In the Prisoner's Dilemma example in Section~\ref{sec:intro}, even if we knew the true type from the onset, a depth of $h = 3$ would predict higher average payoffs for the expert $D$, while $C$ is more profitable in the long-term. Thus, a higher $h$ would be needed.

Likewise, $U^*$ may be inaccurate if the true type of player $j$ is not included in $\Theta_j^*$. However, in this case, the confidence $C^t$ should adjust the portion of $U^*$ used in the mix. As we show in our experiments, this can ensure that the expert algorithm only suffers minor (or no) degradations in its performance. Moreover, if there are types in $\Theta_j^*$ which are \emph{similar} to the true type, in the sense that they assign similar probabilities to actions, then the accuracy of $U^*$ often remains at a level which is proportional to the similarity.




	\subsection{Total Payoffs} \label{sec:total}

\ehba\ can be applied to any expert algorithm which considers the average payoffs that experts yielded in the past, i.e. $\bar{U}$. However, there are some expert algorithms that consider \emph{total} rather than average payoffs, e.g. Hedge \cite{fs1995} and Exp3 \cite{acfs1995}.

One way in which \ehba\ can be applied to such expert algorithms is to modify them to use average rather than total payoffs. This is possible because, when payoffs are observed iteratively, the total payoff can be mapped into the average payoff. However, it is likely that such a modification would invalidate the performance guarantees of the expert algorithm. For example, Hedge and Exp3 will not converge to the best expert if they are modified to use average payoffs.

Another way to apply \ehba\ to such expert algorithms is to modify the definitions of \ehba\ to use total rather than average payoffs. That is, we redefine $\bar{U}$ to be the \emph{total} payoff the expert yielded when following their recommendations, and we compute $U^*$ using \eqref{eq:EQ}/\eqref{eq:Qmod} but without dividing by $h$. This is the approach we have chosen in our experiments.

However, there are two potential complications with the latter approach, both of which are due to the fact that $\bar{U}$ is the total payoff over (up to) $t$ steps while $U^*$ is the total payoff over $h$ steps. Firstly, this means that there is no notion of accuracy of $\bar{U}$ and $U^*$, hence our correctness claim in Section~\ref{sec:guaran} regarding the accuracy of the mixing does not hold anymore. Secondly, as the game proceeds with ever increasing $t$, the impact of $U^*$ in the mix will diminish over time for $C^t < 1$. Nonetheless, as we show in our experiments, if $C^t$ converges to 1, then \ehba\ can still significantly improve the performance of the expert algorithm.

	\section{Experiments} \label{sec:exp}

We conducted experiments in a comprehensive set of repeated matrix games, comparing the performance of several well-known expert algorithms with and without \ehba.

		\subsection{Games}

We used a comprehensive set of benchmark games introduced by \citp{rg1966}, which consists of 78 repeated $2 \times 2$ matrix games (i.e. 2 players with 2 actions). The games are \emph{strictly ordinal}, meaning that each player ranks each of the 4 possible outcomes from 1 (least preferred) to 4 (most preferred), and no two outcomes have the same rank. Furthermore, the games are \emph{distinct} in the sense that no game can be obtained by transformation of any other game, which includes interchanging the rows, columns, and players (and any combination thereof) in the payoff matrix of the game.

The games can be grouped into 21 \emph{no-conflict} games and 57 \emph{conflict} games. In a no-conflict game, the two players have the same most preferred outcome, and so it is relatively easy to arrive at a solution that is best for both players. In a conflict game, the players disagree on the best outcome, hence they will have to find some form of a compromise.

		\subsection{Experts \& Types} \label{sec:types}

We used three automatic methods to generate parameterised sets of experts and types for a given game. The generated policies cover a reasonable spectrum of adaptive behaviours, including deterministic (CDT), randomised (CNN), and hybrid (LFT) policies. All parameter settings can be found in the appendix \cite{acr2015ehbaapp}.

			\subsubsection{Leader-Follower-Trigger Agents (LFT)} \label{sec:lfg}

\citp{c2014} described a method to automatically generate sets of ``leader'' and ``follower'' agents that seek to play specific sequences of joint actions, called ``target solutions''. A leader agent plays its part of the target solution as long as the other player does. If the other player deviates, the leader agent punishes the player by playing a minimax strategy. The follower agent is similar except that it does not punish. Rather, if the other player deviates, the follower agent randomly resets its position within the target solution and continues play as usual. We augmented this set by a trigger agent which is similar to the leader and follower agents, except that it plays its maximin strategy indefinitely once the other player deviates.

			\subsubsection{Co-Evolved Decision Trees (CDT)}

We used genetic programming \cite{k1992} to automatically breed sets of decision trees. A decision tree takes as input the past $n$ actions of the other player (in our case, $n = 3$) and deterministically returns an action to be played in response. The breeding process is co-evolutional, meaning that two pools of trees are bred concurrently (one for each player). In each evolution, a random selection of the trees for player 1 is evaluated against a random selection of the trees for player 2. The fitness criterion includes the payoffs generated by a tree as well as its dissimilarity to other trees in the same pool. This was done to encourage a more diverse breeding of trees, as otherwise the trees tend to become very similar or identical.

			\subsubsection{Co-Evolved Neural Networks (CNN)}

We used a string-based genetic algorithm \cite{h1975} to breed sets of artificial neural networks. The process is basically the same as the one used for decision trees. However, the difference is that artificial neural networks can learn to play stochastic strategies while decision trees always play deterministic strategies. Our networks consist of one input layer with 4 nodes (one for each of the two previous actions of both players), a hidden layer with 5 nodes, and an output layer with 1 node. The node in the output layer specifies the probability of choosing action 1 (and, since we play $2 \times 2$ games, of action 2). All nodes use a sigmoidal threshold function and are fully connected to the nodes in the next layer.

		\subsection{Expert Algorithms}

The following expert algorithms were used: UCB1 \cite{acf2002}, EEE \cite{fm2004}, S \cite{kmrv1998}, Hedge \cite{fs1995}, and Exp3 \cite{acfs1995}.

For EEE, we used the parameter settings specified in \cite{fpm2003}. S was implemented as specified in Appendix~A in \cite{c2014}, using the same parameter settings. For Hedge, we used the modified version provided in Section~3 in \cite{acfs1995}. Both Hedge and Exp3 used $\eta = 0.1$, and Exp3 used $\gamma = 0.1$.

As discussed in Section~\ref{sec:total}, Hedge and Exp3 are based on \emph{total} rather than average payoffs, hence we adapted \ehba\ as specified in Section~\ref{sec:total}. That is, the variable $G_i$ in Hedge and Exp3 (cf. \cit{acfs1995}) was used in place of $U^*$. (Note that, in Hedge, $G_i$ is the total payoff of \emph{each} expert's recommendations, not just of those which we followed.) In addition, we applied a ``booster'' exponent $b$ to $U^*$ (i.e. $(U^*)^b$) to magnify the differences between experts. We used $b = 3$.

		\subsection{Experimental Procedure}

We performed identical experiments for every expert/type generation method described in Section~\ref{sec:types}. Each of the 78 games was played 10 times using different random seeds, where each play lasted 5,000 rounds (this was unknown to the players to avoid ``end-game'' effects).

In each play, we randomly generated 5 unique experts for player 1 (controlled by \ehba) and 5 unique types for player 2, and provided them to \ehba\ as the sets $\Phi_1$ and $\Theta^*_2$, respectively. \ehba\ used uniform prior beliefs and a planning horizon of $h=5$, which constituted a good trade-off between computation time and accuracy in our experiments.

Every play was repeated in two modes: one in which player 2 was controlled by a randomly generated type, and one in which it was controlled by a fictitious player \cite{b1951}. We used a fictitious player because it explicitly tries to learn the behaviour of \ehba. While the generated types are adaptive as well, they do not create models of \ehba's behaviour.

Finally, to isolate the effects of knowing the true type of player 2, we performed each experiment once for the case in which the true type of player 2 was included in $\Theta^*_2$ and once for the case in which it was not ($|\Theta^*_2| = 5$ in both cases).

		\subsection{Results}

\begin{figure*}[h]
	\vspace{-10pt}
	\centering
	\subfloat[LFT -- RT -- No-Conflict]{\label{fig:inc-1}\includegraphics[height=\h]{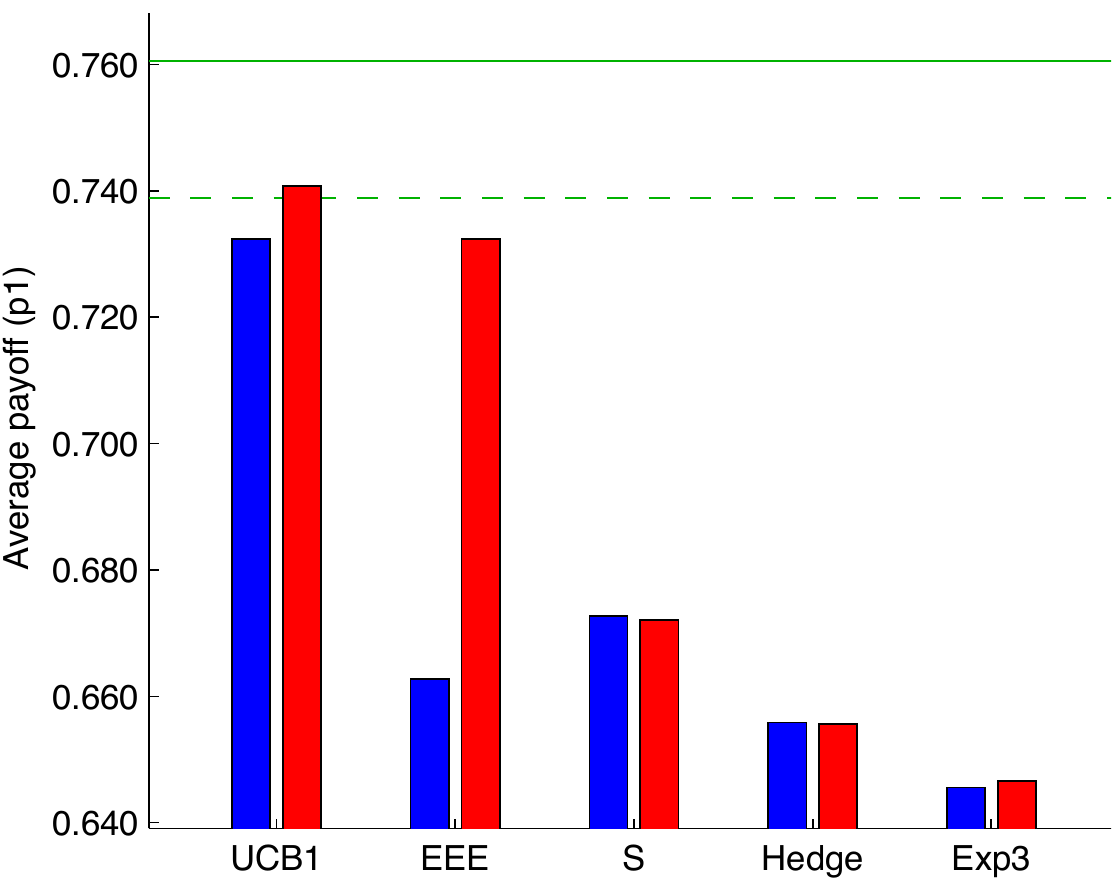}}\hspace{\hs}
	\subfloat[CDT -- RT -- Conflict]{\label{fig:inc-2}\includegraphics[height=\h]{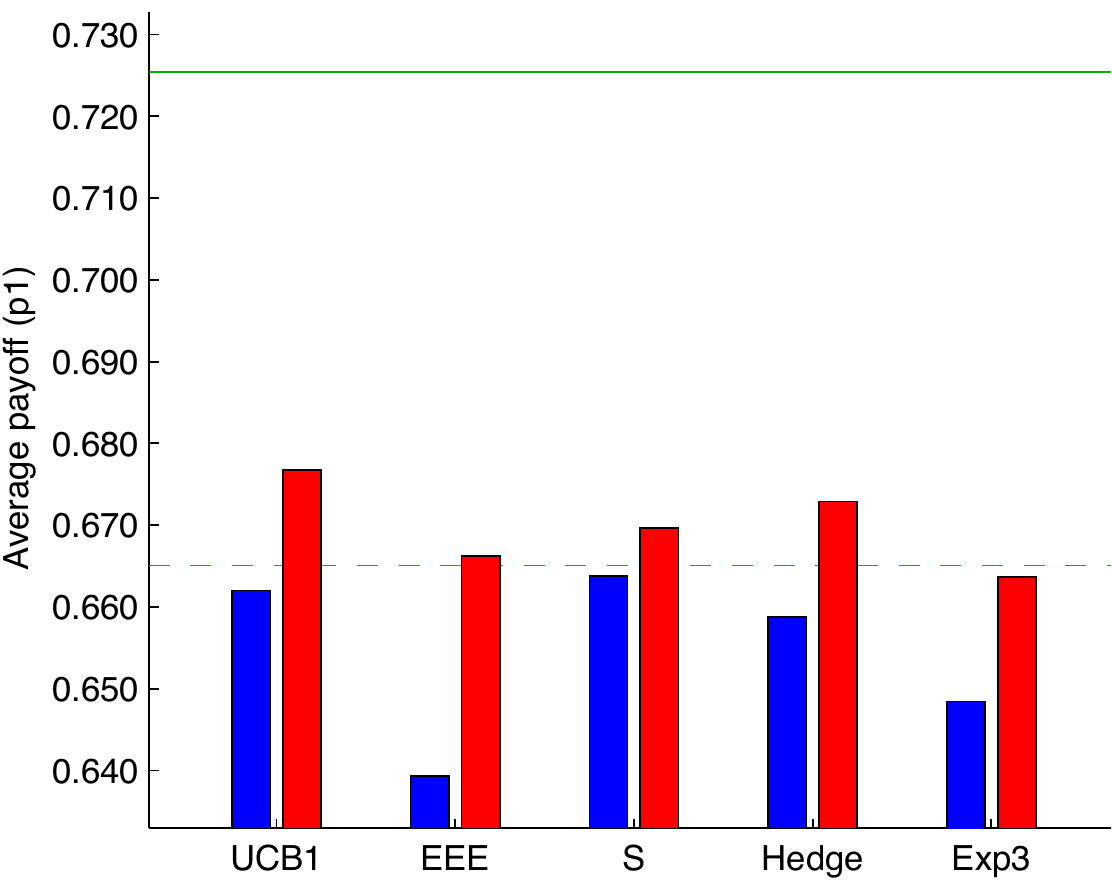}}\hspace{\hs}
	\subfloat[CNN -- RT -- No-Conflict]{\includegraphics[height=\h]{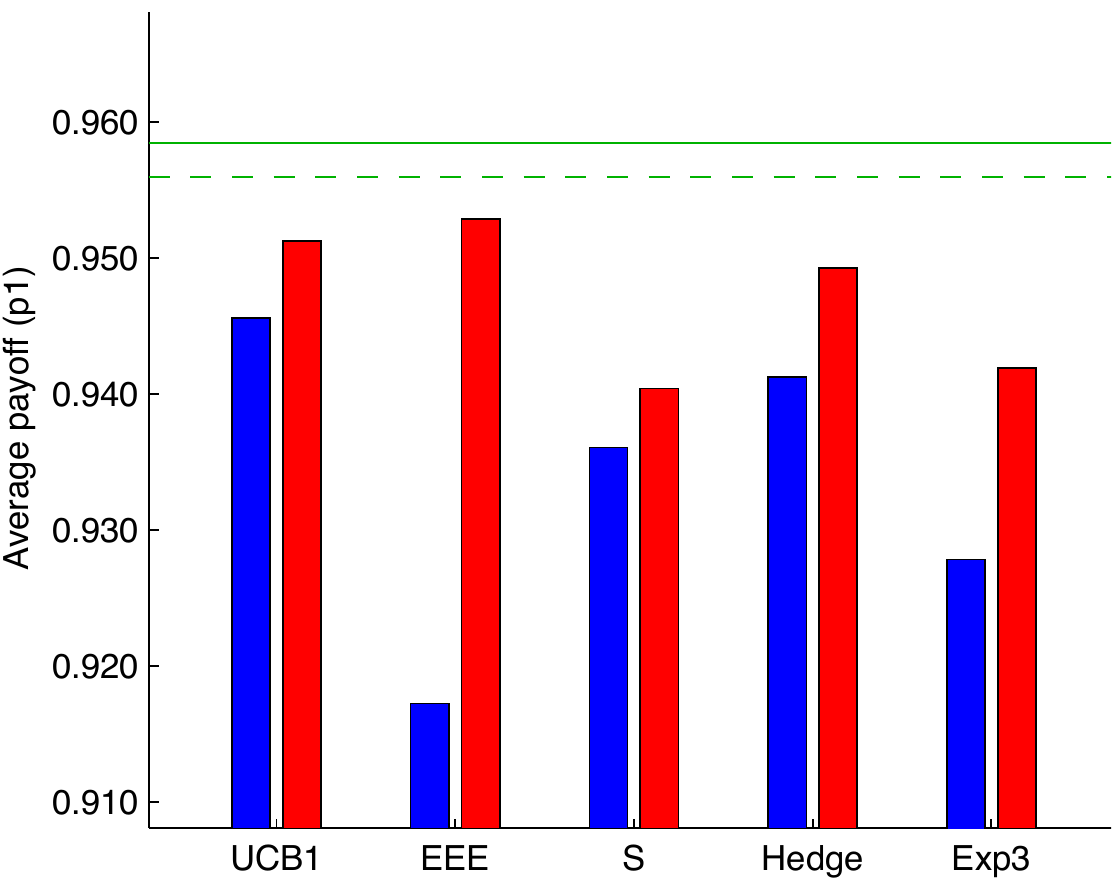}}\\
	\subfloat[LFT -- FP -- Conflict]{\includegraphics[height=\h]{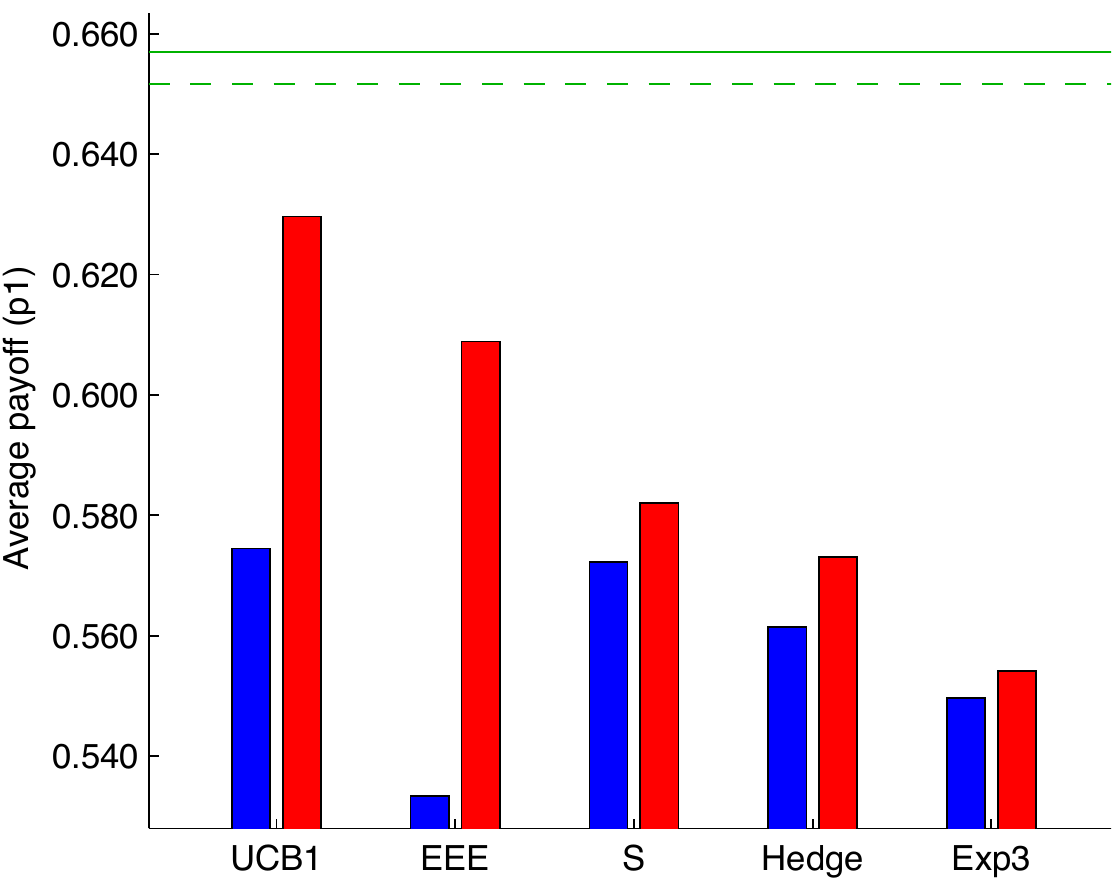}}\hspace{\hs}
	\subfloat[CDT -- FP -- Conflict]{\includegraphics[height=\h]{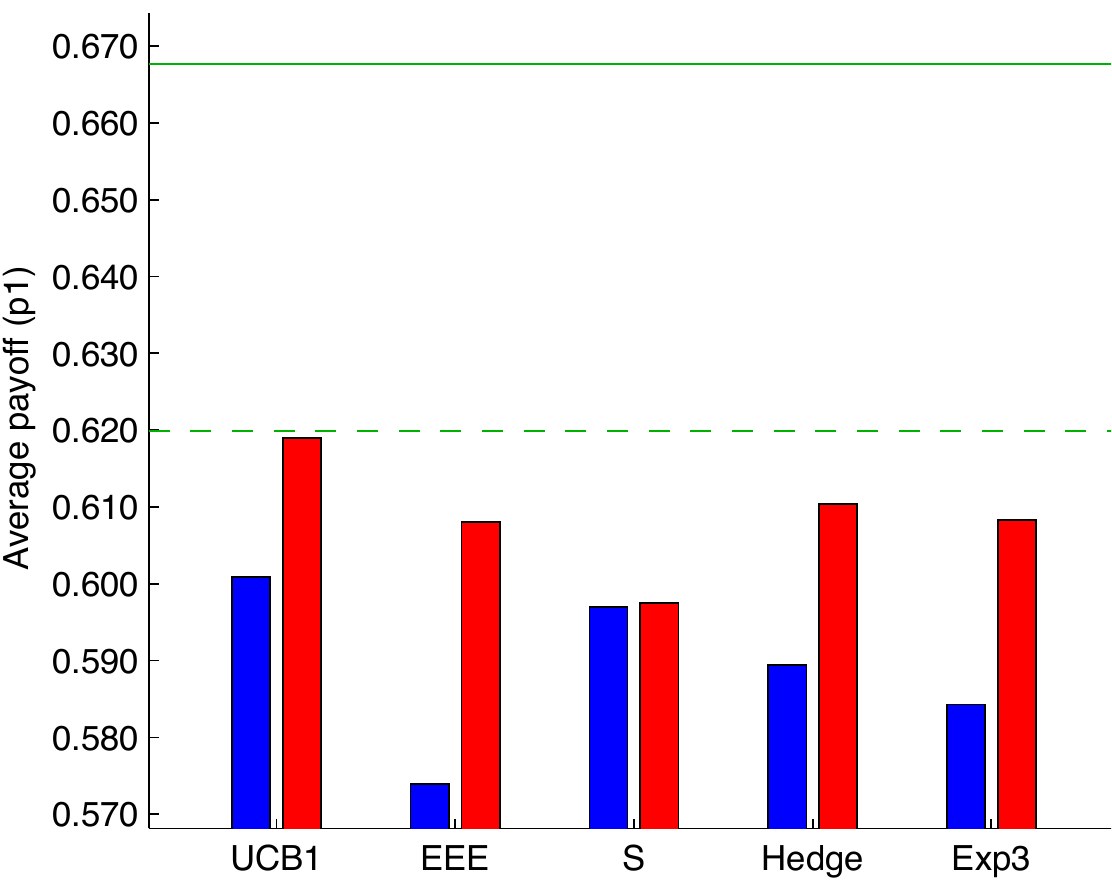}}\hspace{\hs}
	\subfloat[CNN -- FP -- Conflict]{\includegraphics[height=\h]{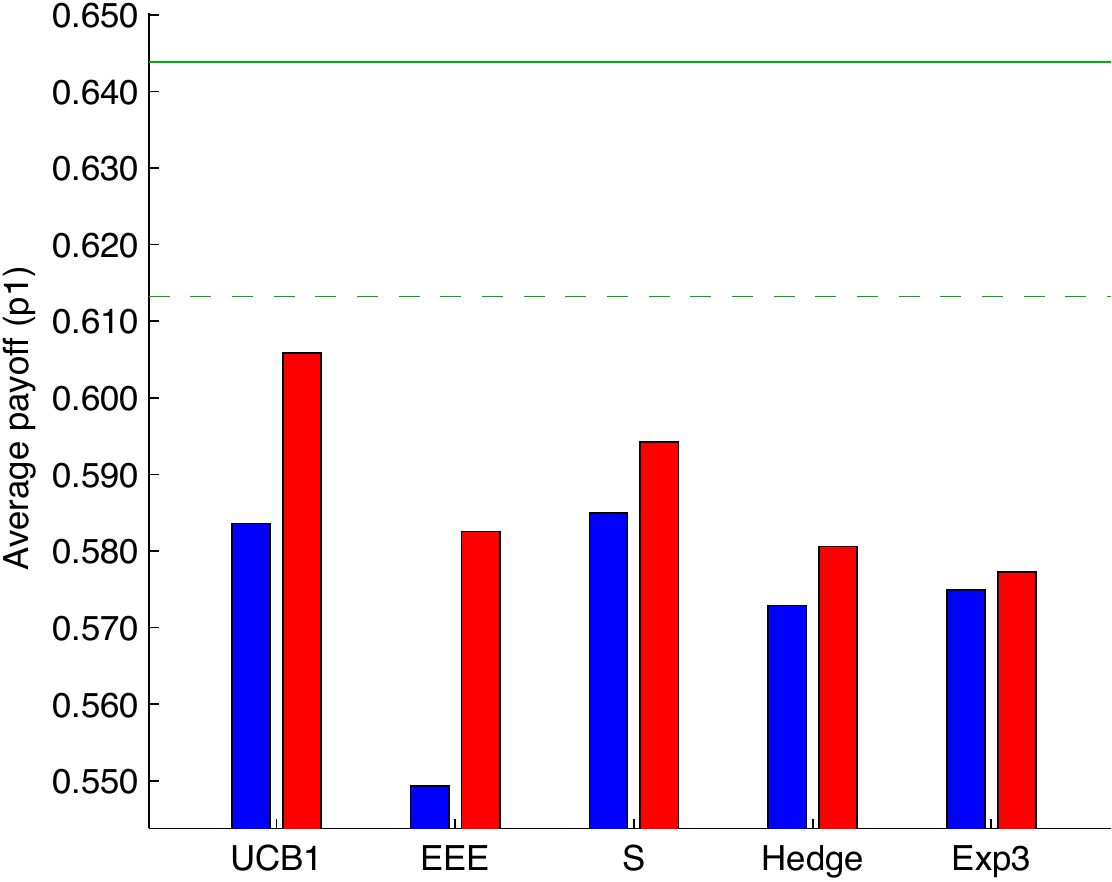}}
	\caption{\textbf{True type of player 2 included in $\Theta_2^*$.} X--Y--Z format means that experts and types were generated by X, player 2 was controlled by Y, and results are shown for Z games. RT denotes random type and FP denotes fictitious player.}
	\label{fig:inc}
	
	\vspace{17pt}
\end{figure*}

Figures \ref{fig:inc} and \ref{fig:noinc} show a selection of results from a variety of scenarios (all results are given in the appendix document). Since most of the tested expert algorithms assume payoffs in the interval $[0,1]$, we normalised the payoffs from $\left\{ 1,2,3,4 \right\}$ to $\left\{ 0,\frac{1}{3},\frac{2}{3},1 \right\}$. The solid and dashed lines show the average payoffs of HBA (using the same parameters and types as \ehba) and the best expert in each play, respectively. In the following, all 'significance' statements are based on paired two-sided t-tests with a significance level of 5\%.

Figure~\ref{fig:inc} shows results for the case in which the true type of player 2 was included in $\Theta_2^*$. As the plots show, \ehba\ was able to significantly improve the performance of the expert algorithms. This was observed in all constellations of games, experts/types, and opponents. UCB1 and EEE benefited the most from \ehba, with improvements of up to 10\% and 15\%, respectively. The improvements of the other algorithms were less substantial but still significant. Note that HBA performed the best in all experiments because it computes best-response actions at each point in time, whereas the expert algorithms (with and without \ehba) must choose from a set of pre-defined expert policies that may not be optimal.

Another interesting observation is that \ehba\ often meant the difference between performing worse and better than the best expert (e.g. Figures~\ref{fig:inc-1} and \ref{fig:inc-2}). This is precisely a result of the fact that the tested expert algorithms rely heavily on the past performance of experts (as discussed in Section~\ref{sec:intro}), while \ehba\ also considers the future performance of experts at any given point. This allowed the modified expert algorithms to switch effectively between experts, which on average resulted in higher payoffs than persistently following any single expert.

It is important to note that the performance enhancements of \ehba\ came at an increased computational cost. For instance, using a planning depth of $h = 5$ and the CDT types, the modified expert algorithms required roughly 10 times more time than the original algorithms. (However, we note that our implementation was not optimised for speed.) Therefore, whether or not the performance improvements are worth the additional costs is an important aspect that should be taken into account. Nonetheless, as discussed previously, the computational costs can often be reduced drastically by using efficient Monte-Carlo Tree Search methods.

Figure~\ref{fig:noinc} shows results for the case in which the true type of player 2 was \emph{not} included in $\Theta_2^*$. Here, we observe that the modified algorithms performed similarly to the original algorithms, and substantially better than HBA (which often ended up playing randomly). This was due to the confidence $C^t$, which decreased quickly and remained low in many cases. Interestingly, there are cases in which the expert algorithms still benefited from \ehba, especially with CDT and CNN in no-conflict games. In these cases, the set $\Theta_2^*$ included a type that was similar to the true type (in the sense that they assigned similar probabilities to actions) so that \ehba\ and HBA were still able to make useful predictions.

Finally, we note that, in some cases, Hedge and Exp3 performed significantly worse when combined with \ehba. We found that this was mostly due to the problems described Section~\ref{sec:total}. Thus, while \ehba\ works well in combination with expert algorithms that use average payoffs, it may not be optimal for expert algorithms that use total payoffs

	\section{Conclusion} \label{sec:conc}

Past research has studied two methods to utilise pre-defined policy sets in repeated interactions: as \emph{experts}, to dictate our own actions, and as \emph{types}, to characterise the behaviour of other agents. The contribution in this work is to bring these complementary views together, with the goal of combining their strengths and alleviating their weaknesses.

We have done so in the form of a meta-algorithm, \ehba, which can be applied to any expert algorithm that considers the average (or total) payoff an expert yielded in the past. \ehba\ mixes these observed payoffs with a predicted future payoff for each expert, using the type-based approach. Our experiments show that, if the true (or a similar) type of the opponent is known, then \ehba\ can significantly improve the performance of expert algorithms, while in all other cases it performs similarly to the original expert algorithm.

At the core of \ehba\ is the confidence, $C^t$, which is used to regulate the mixing. We provide an intuitive definition of confidence which is simple to compute and works well in practice. However, we believe that other useful definitions of confidence exist, and in this sense we view \ehba\ as a \emph{family} of meta-algorithms. An important step in establishing this idea will be to develop a theory around this notion of confidence, similar to the theories on regret in expert algorithms and the theories on prior and posterior beliefs in the type-based approach.


\clearpage

\begin{figure*}[h]
	\vspace{-10pt}
	\centering
	\subfloat[LFT -- RT -- No-Conflict]{\includegraphics[height=\h]{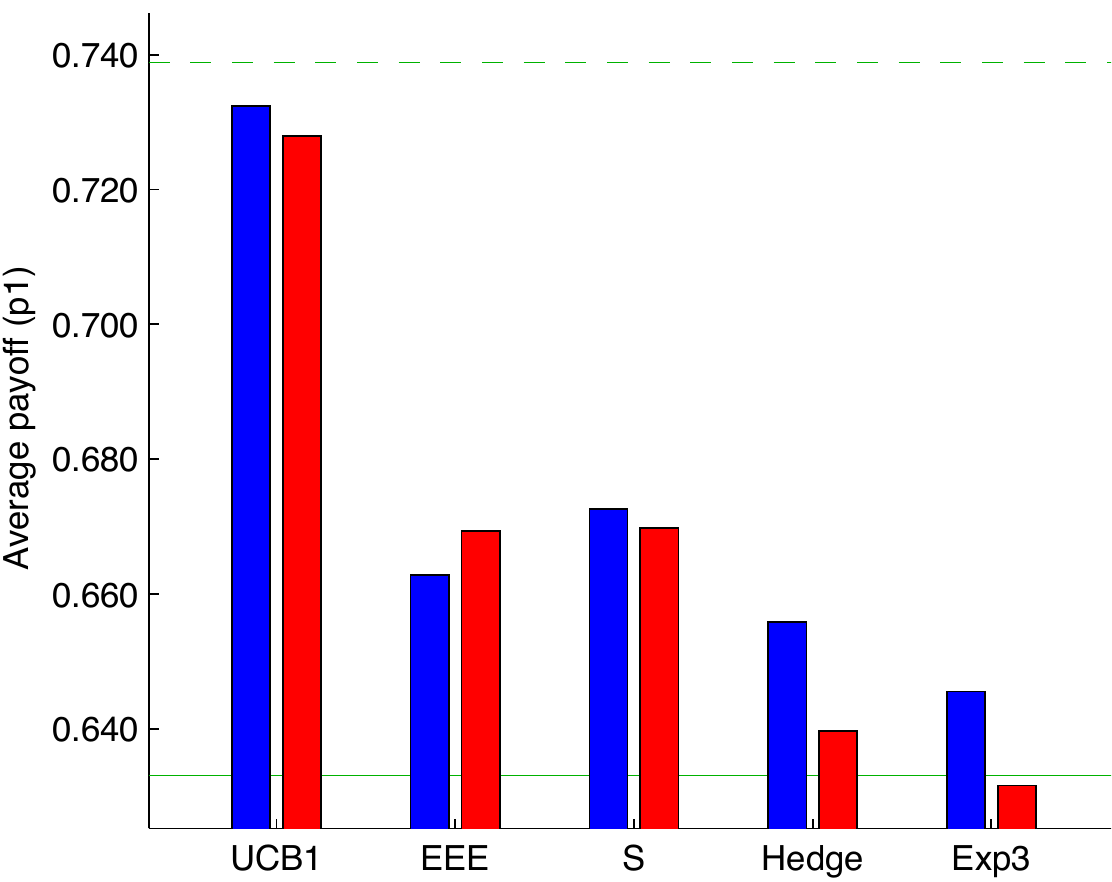}}\hspace{\hs}
	\subfloat[CDT -- RT -- Conflict]{\includegraphics[height=\h]{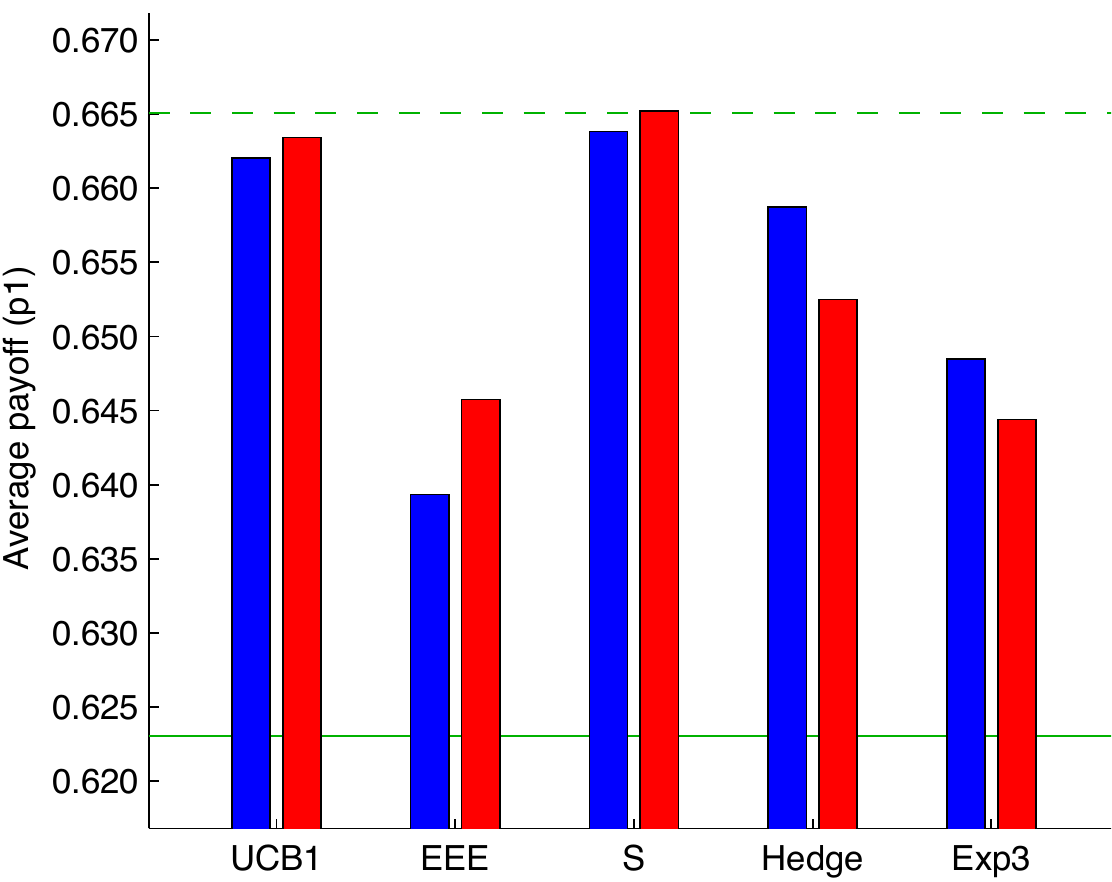}}\hspace{\hs}
	\subfloat[CNN -- RT -- No-Conflict]{\includegraphics[height=\h]{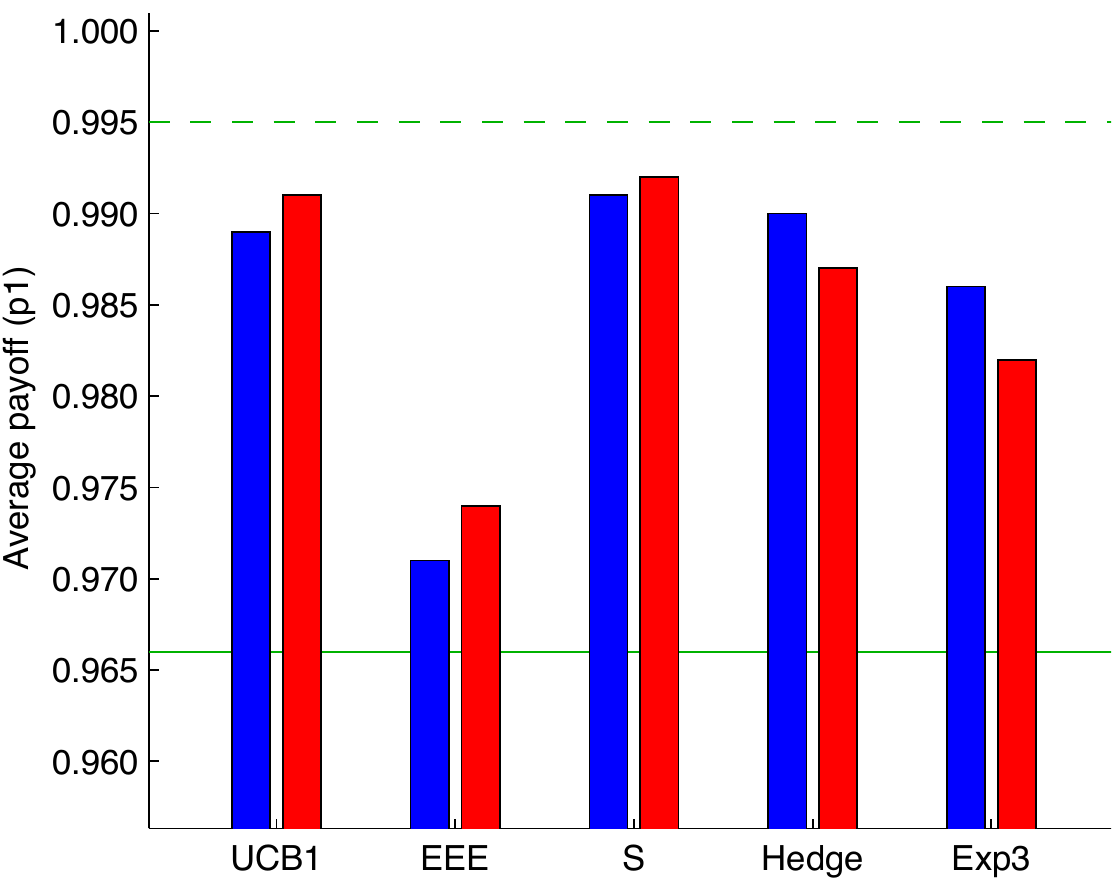}}\\
	\subfloat[LFT -- FP -- Conflict]{\includegraphics[height=\h]{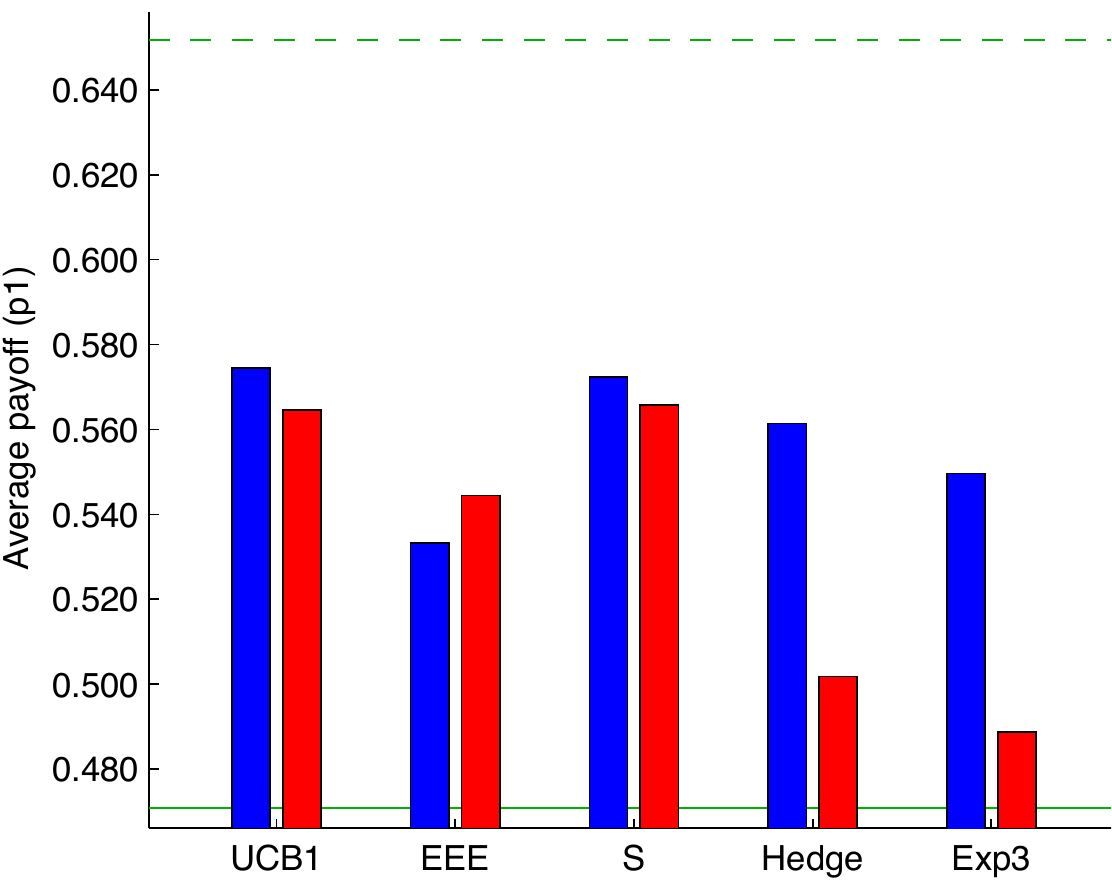}}\hspace{\hs}
	\subfloat[CDT -- FP -- Conflict]{\includegraphics[height=\h]{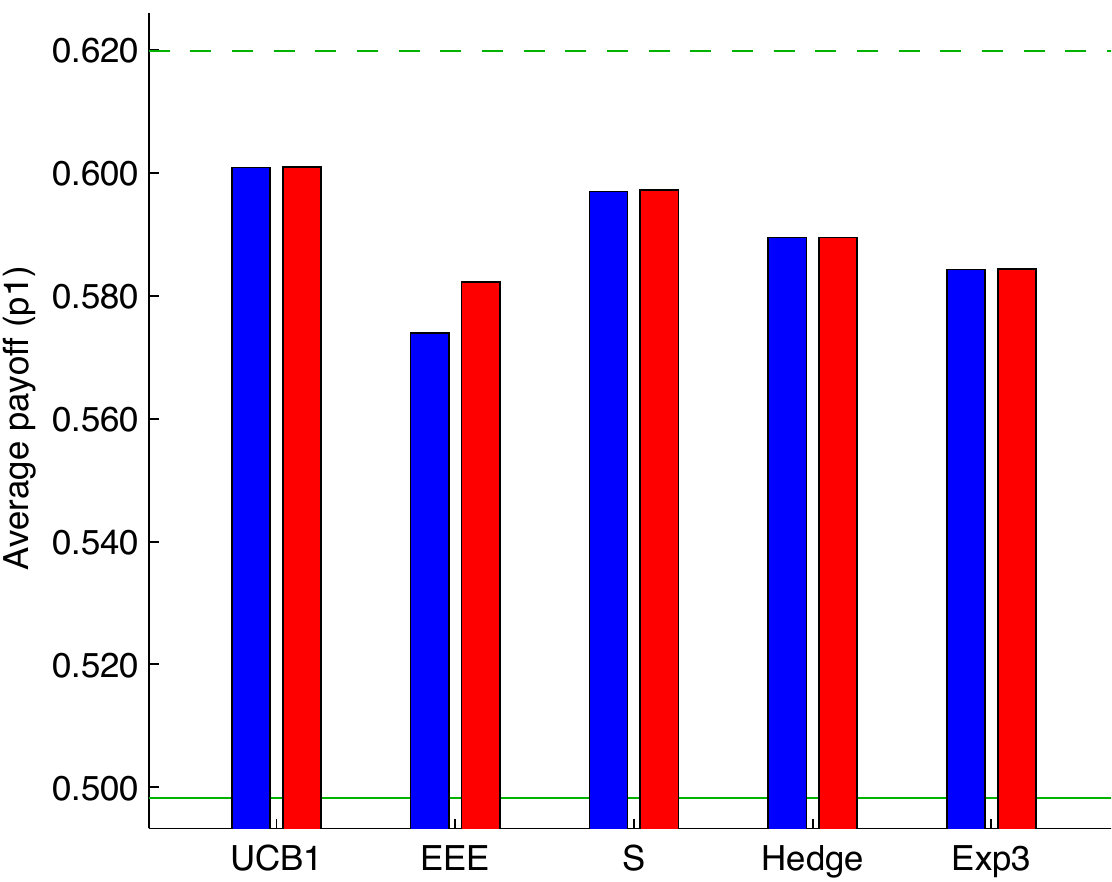}}\hspace{\hs}
	\subfloat[CNN -- FP -- Conflict]{\includegraphics[height=\h]{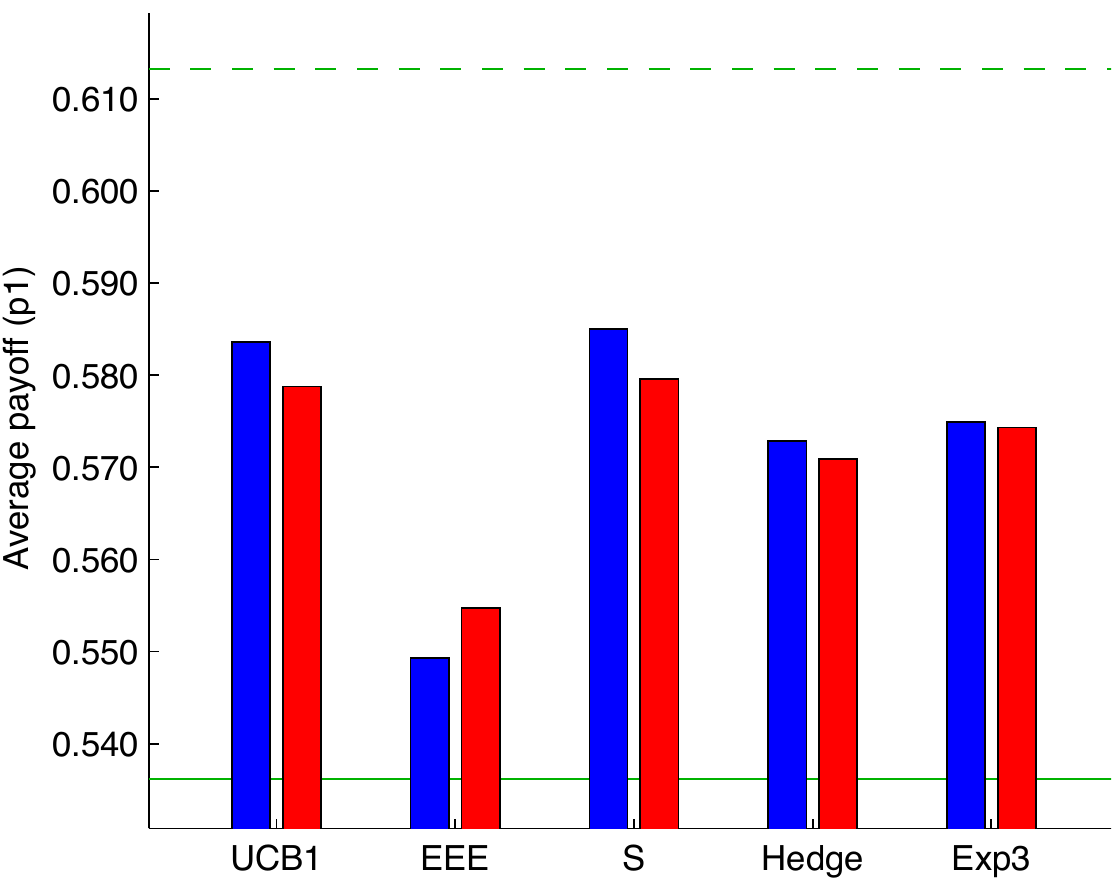}}
	\caption{\textbf{True type of player 2 \emph{not} included in $\Theta_2^*$.} X--Y--Z format means that experts and types were generated by X, player 2 was controlled by Y, and results are shown for Z games. RT denotes random type and FP denotes fictitious player.}
	\label{fig:noinc}
\end{figure*}

\begin{textblock}{20}(49.1,20.3)
	\includegraphics[height=0.048\textheight]{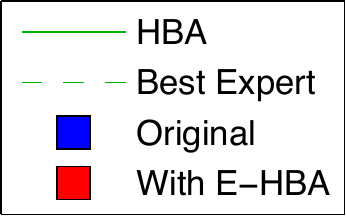}
\end{textblock}

\begin{textblock}{20}(49.1,54.0)
	\includegraphics[height=0.048\textheight]{legend.pdf}
\end{textblock}

	\clearpage
	\bibliographystyle{aaai}
	\bibliography{aaai15_ehba}

\end{document}